\documentclass[letterpaper, 10 pt, conference]{ieeeconf}  

\IEEEoverridecommandlockouts                              

\makeatletter
\let\NAT@parse\undefined
\makeatother

\usepackage{graphics} 
\usepackage{epsfig} 
\usepackage{amsmath} 

\usepackage{color}
\usepackage{url}
\usepackage{hyperref}
\usepackage{amsmath}
\usepackage{relsize}
\usepackage{graphicx}
\usepackage{amssymb}
\usepackage{booktabs}
\usepackage{multirow}
\usepackage{rotating}
\definecolor{jeremy_color}{rgb}{0,.7,.7}
\definecolor{patrick_color}{rgb}{.6,.4,.05}
\definecolor{charlie_color}{rgb}{0,0,0.8}
\definecolor{cody_color}{rgb}{0.75,0.25,0.0}

\newcommand{\comment}[1]{} 

\newcommand{\method}[0]{ViPER}
\newcommand{\network}[0]{ViPER}

\title{\LARGE \bf Visual Contact Pressure Estimation for Grippers in the Wild}

\author{Jeremy A. Collins$^{1}$, Cody Houff$^{1}$, Patrick Grady$^{1}$, Charles C. Kemp$^{1}$
\footnotemark{}
}

\newenvironment{first_caption}
  {\par\footnotesize}
  {\par\addvspace{\bigskipamount}}

\begin{document}

\twocolumn[{%
\renewcommand\twocolumn[1][]{#1}%

\maketitle
\thispagestyle{empty}
\pagestyle{empty}

\begin{center}
\centering
\vspace{-2mm}
\includegraphics[width=1.0\linewidth]{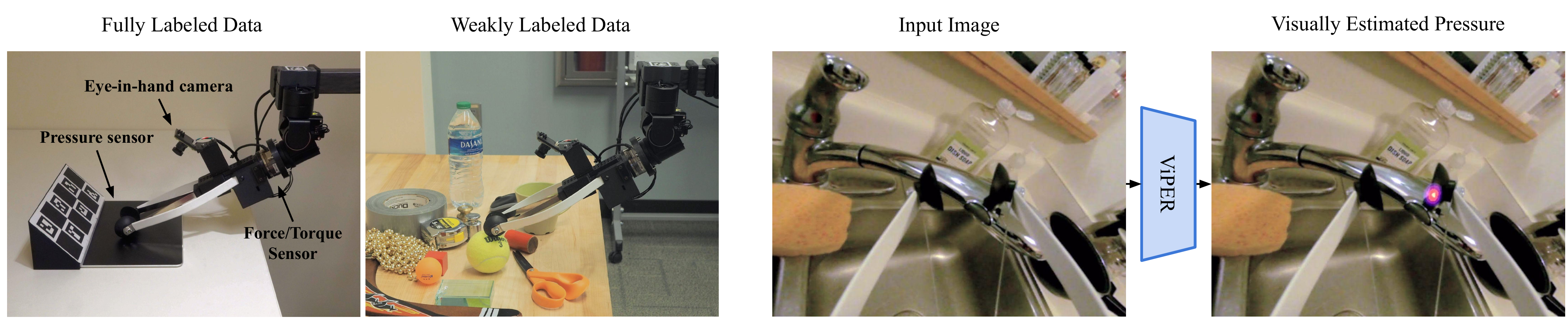}
\end{center}
\vspace{0mm}
\begin{first_caption}
Fig. 1. \quad \textbf{Fully Labeled Data:} In controlled settings, we collect training data consisting of images, contact pressure, forces, and torques. \textbf{Weakly Labeled Data:} In uncontrolled settings, we collect training data consisting of images, forces, and torques. \textbf{Our Model:} We use fully labeled and weakly labeled data to train \network{}. During test time, \network{} takes only an RGB image from an eye-in-hand camera as input and outputs a contact pressure estimation as well as force and torque estimates. \network{} outperforms prior methods, enables precision manipulation in clutter, and provides accurate estimates in unseen conditions.
\end{first_caption}
\vspace{-1mm}
}]
\setcounter{figure}{1}  
\setcounter{footnote}{1}

\footnotetext{Jeremy A. Collins, Cody Houff, Patrick Grady, and Charles C. Kemp are with the Institute for Robotics and Intelligent Machines at the Georgia Institute of Technology (GT). This work was supported in part by NSF Award \# 2024444 and AI-CARING Award \# 2112633. Charles C. Kemp is an associate professor at GT. He also owns equity in and works part-time for Hello Robot Inc., which sells the Stretch RE1. He receives royalties from GT for sales of the Stretch RE1.}

\begin{abstract}

Sensing contact pressure applied by a gripper can benefit autonomous and teleoperated robotic manipulation, but adding tactile sensors to a gripper's surface can be difficult or impractical. If a gripper visibly deforms, contact pressure can be visually estimated using images from an external camera that observes the gripper. While researchers have demonstrated this capability in controlled laboratory settings, prior work has not addressed challenges associated with visual pressure estimation in the wild, where lighting, surfaces, and other factors vary widely. We present a model and associated methods that enable visual pressure estimation under widely varying conditions. Our model, Visual Pressure Estimation for Robots (ViPER), takes an image from an eye-in-hand camera as input and outputs an image representing the pressure applied by a soft gripper. Our key insight is that force/torque sensing can be used as a weak label to efficiently collect training data in settings where pressure measurements would be difficult to obtain. When trained on this weakly labeled data combined with fully labeled data that includes pressure measurements, ViPER outperforms prior methods, enables precision manipulation in cluttered settings, and provides accurate estimates for unseen conditions relevant to in-home use.

\end{abstract}

\section{Introduction}


By sensing contact pressure, a robot can actively control where its gripper makes contact and the pressure it applies. Yet high-resolution pressure sensors can be infeasible to add to a gripper due to factors such as surface curvature, material compliance, and wear during manipulation.  

If a gripper visibly deforms when applying pressure, visual pressure estimation becomes an option. Visually estimating contact pressure from observations of an unmodified gripper avoids challenges associated with mounting pressure sensors to contact surfaces, since cameras are a mature sensing technology that can observe large regions from a distance. Prior work has shown that contact pressure can be estimated from a single RGB image due to visible deformation of a soft gripper \cite{vpec}. These visual pressure estimates can enable image-based control. For example, a gripper can pick up small objects using visual servoing to control pressure with respect to a target. 



A significant challenge for visual pressure estimation is achieving acceptable performance in the wild. Images of a soft gripper applying the same contact pressure can vary dramatically due to factors such as lighting and the appearance of surfaces. 

Data-driven methods typically depend on training data representing the types of variation that could be encountered. For example, the model in \cite{vpec} was trained and tested in a controlled setting with fixed cameras observing an uncluttered, smooth white surface, and our results (Table \ref{tab:main_results}) suggest it would perform poorly on textured surfaces.


Mounting pressure sensors to the gripper or the environment is difficult and alters the appearance of surfaces as well as their contact mechanics. Using physics modeling and simulation presents an alternative approach, but accurately simulating gripper deformation and visual phenomena associated with contact pressure is challenging. 


We present Visual Pressure Estimation for Robots (ViPER), a novel method that estimates contact pressure given a single RGB image from an eye-in-hand camera. Our key insight is that the forces and torques from a 6-axis wrist-mounted force/torque sensor can serve as a \textit{weak label} for pressure estimation, enabling diverse training examples to be efficiently collected in the wild. Unlike a pressure sensor, a force/torque sensor does not need to be mounted on the contact surfaces of the gripper or the environment, and can be placed behind the camera, so as not to interfere with images. The force/torque sensor is only used for training, and is not needed at test time.

In a controlled setting, we use a pressure-sensing array and a wrist-mounted force/torque sensor to collect images with associated contact pressure measurements, forces and torques. We refer to this as \textit{fully labeled} data. In uncontrolled settings, we collect images with only associated forces and torques, which we refer to as \textit{weakly labeled} data. 

By training our model on both fully labeled and weakly labeled data, the model learns to estimate contact pressure in uncontrolled settings. Although the mapping from contact pressure to forces and torques is many to one, forces and torques provide a strong hint during training as to which contact pressure patterns would be appropriate. 

In summary, we make three main contributions:
\begin{itemize}
    \item \textbf{Visual Estimation of Contact Pressure in the Wild:} We present \network{}, a neural network capable of visually estimating contact pressure for a soft gripper in the wild using only an RGB image from an eye-in-hand camera. \network{} enables precision grasping of small objects (e.g., a paperclip) on unseen cluttered surfaces. 
    \item \textbf{Force/Torque Sensing as a Weak Label for Contact Pressure:} We show that forces and torques are an effective weak label for training a model to estimate contact pressure. 
    \item \textbf{Open Releases:} Our code, data, models, and custom hardware can be found at \href{https://github.com/Healthcare-Robotics/ViPER}{github.com/Healthcare-Robotics/ViPER}.

\end{itemize}

\begin{figure}[t]
  \centering
  \includegraphics[width=1\linewidth]{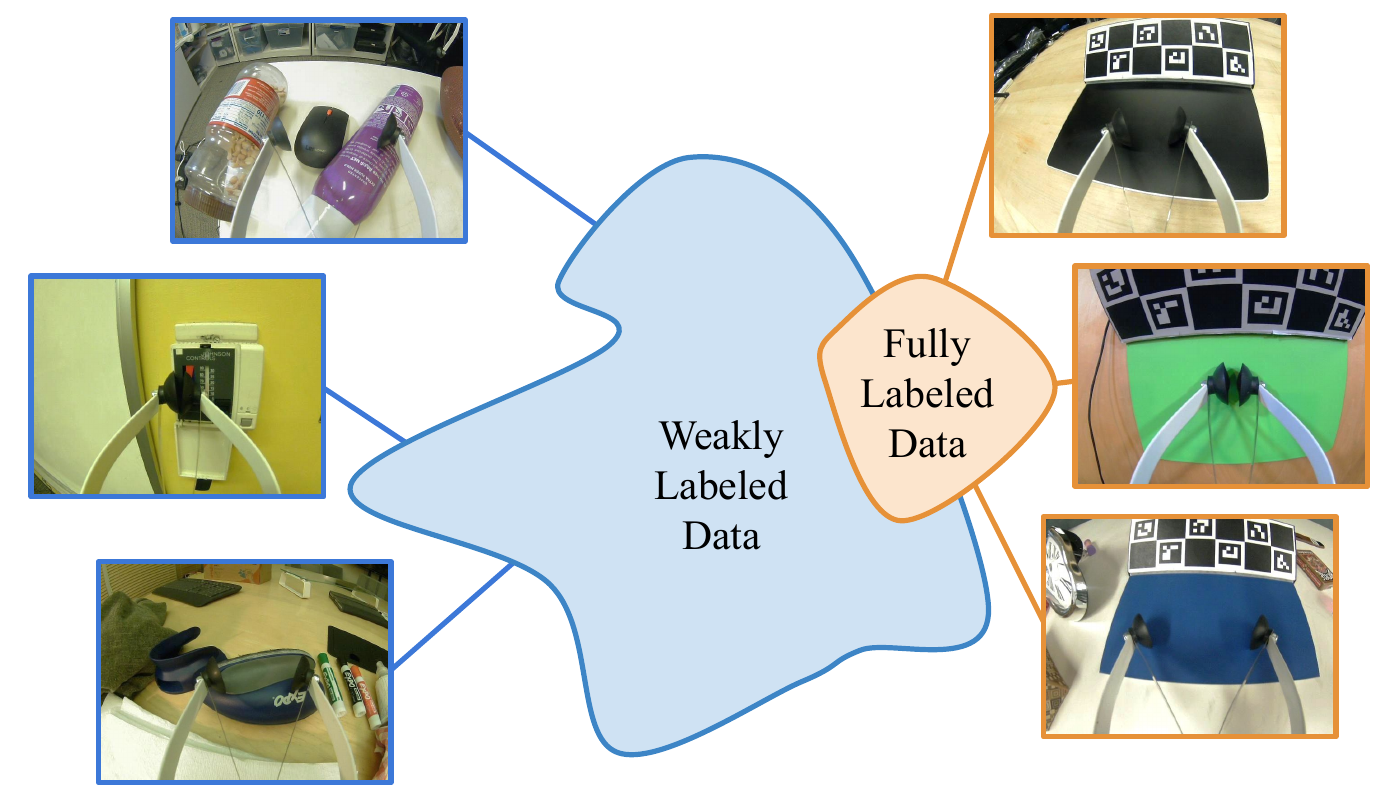}
  \caption{Images captured in the fully labeled setting always contain a large, flat pressure sensor and a fiducial board. Alternatively, weakly labeled data can be collected in realistic diverse scenarios, such as touching objects, buttons, curved surfaces, and compliant surfaces. We demonstrate that weakly labeled data improves estimation performance and enables manipulation in diverse settings.}
  \label{fig:domain_figure}
\end{figure}

\begin{figure*}[t]
  \centering
  \vspace{2mm}
  \includegraphics[width=0.85\linewidth]{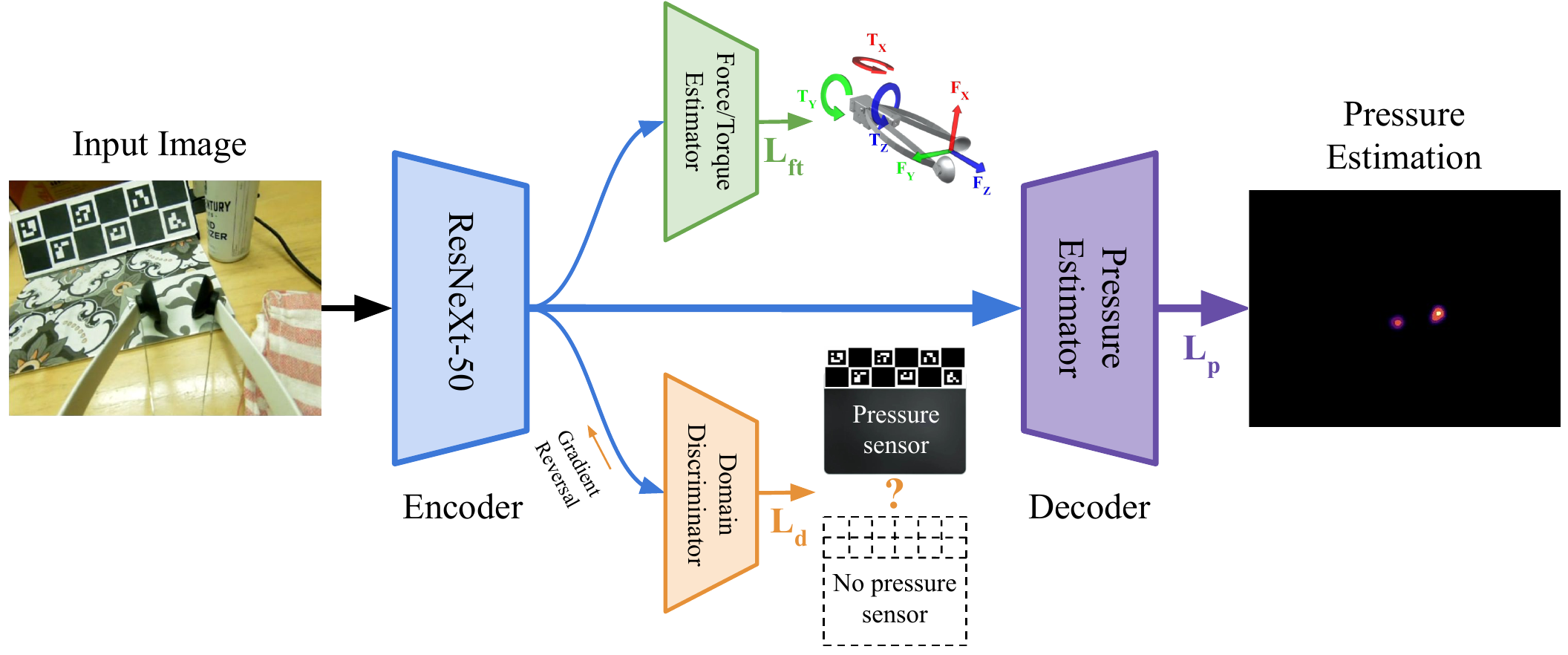}
  \vspace{-2mm}
  \caption{\network{} architecture. An input image first passes through a convolutional image encoder. The resulting embeddings are used as input to a force/torque predictor, an adversarially trained domain discriminator, and a pressure estimator. The pressure estimator outputs a contact pressure image.}
  \label{fig:architecture}
\end{figure*}

\section{Related Work}

\subsection{Visual Estimation of Contact and Pressure}

A variety of methods have been proposed to estimate contact between robots and the environment from vision. Many techniques rely on compliance to determine the location and magnitude of force. 

One body of work focuses on stiff tools applying force to compliant objects. For example, surgical robots may estimate forces by sensing the deformation in soft tissue caused by a rigid tool \cite{nazari2021image, kennedy2005vision, noohi2014using, marban2019recurrent, chua2021toward}. Other work uses CNNs to estimate the force applied to household objects \cite{kim2019efficient}. Recent work has also used physics simulations to train deep models that estimate force applied by cloth to stationary rigid objects using point clouds \cite{erickson_visual_haptic_reasoning_2022}.

Another common technique to sense applied force relies on cameras mounted inside the gripper. These methods typically observe a compliant surface on the exterior of the gripper from a camera placed behind the surface within the gripper's interior. For example, force may be inferred with a depth camera \cite{kuppuswamy2020soft}, by tracking marker patterns on the inside of the compliant surface \cite{ward2018tactip, yamaguchi2016combining}, or with photometric stereo \cite{yuan2017gelsight, lambeta2020digit}. An eye-in-hand camera has previously been used to estimate the forces and torques applied to a soft gripper \cite{vfts}.


Our paper builds upon prior work to estimate contact pressure for soft robotic grippers and human hands. \cite{vpec} uses cameras mounted to the environment to observe a soft gripper interacting with a flat pressure sensor with a white surface. The authors collected a small dataset with little diversity, and did not show generalization outside of their limited environment. \cite{grady2023visual} presents a method to use weak labels to prompt human behavior in an uncontrolled setting. In contrast, our work focuses on estimating contact pressure for soft robotic grippers. Our network uses a similar architecture, but we use force/torque sensing to provide weak labels. We also use a novel automated lighting setup to efficiently collect data with lighting variations, achieve high performance across diverse situations, and demonstrate autonomous control in a cluttered setting.


\subsection{Learning with Partial Supervision}

Modern neural networks are often trained in a fully supervised manner, with a large dataset of fully-labeled training examples and a testing dataset drawn from a similar distribution. However, performance decreases on testing data that does not resemble the training data. As a result, researchers have developed several methods to leverage data with either no labels or weak labels.

Unsupervised domain adaptation (UDA) techniques leverage training data with no labels. These techniques often attempt to reduce the distance between embeddings of a source dataset with labels and a target dataset without labels \cite{long2015dan, tzeng2014deepdomainconfusion, long2016unsupervised}. DANN \cite{ganin2016dann} uses a discriminator to identify the domain an embedding came from, then minimizes the difference between embedding distributions via adversarial training.

Weakly labeled data contains partial labels that may be easier to collect than full labels. Even though these labels may not supervise the primary task, they are a stronger form of supervision than UDA. For computer vision tasks, weak labels may improve performance while taking less effort to annotate than full labels \cite{ahn2018learning, chang2020weakly, paul2020domain}.

\section{Visual Pressure Estimation}




In this section, we describe \method{}, a method to leverage both fully labeled and weakly labeled data for visual pressure estimation in diverse environments. 

\subsection{Problem Formulation}

We seek to estimate the pressure, $P$, exerted by a robotic gripper that visibly deforms using a single RGB image, $I$, as input. We learn a function parameterized by a convolutional encoder-decoder network, $\hat{P} = f(I)$, to estimate this pressure. Prior work \cite{vpec} has used supervised learning to obtain this function from a \textit{fully labeled dataset} of image and pressure pairs $D_F=\{I^i_F, P^i_F\}$. However, collecting pressure data requires modifying the environment with physical sensors, changing its visual and tactile properties and limiting the diversity of data that can be collected.

\method{} is able to train on data from natural environments without ground truth pressure data. Instead, we add an additional sensor to the robot which provides a weak label $w$. This weak label is easier to collect than full pressure labels, and provides supervision even when pressure labels are not available. 

In our work, we use measurements from a force/torque sensor as weak labels. The force/torque sensor (Figure 1) measures 3-axis forces and 3-axis torques applied to the gripper. During training, these forces and torques provide information that is indicative of contact and pressure applied by the gripper.

Formally, we collect two types of data for training: a fully labeled dataset $D_F$ which contains images, pressure, and weak labels $D_F=\{I^i_F, P^i_F, w^i_F\}$ and a weakly labeled dataset with only images and weak labels $D_W=\{I^i_W,  w^i_F\}$. 




\subsection{Network Architecture}
We create \network{}, a convolutional encoder-decoder network, to map an RGB image to a pressure estimate in image space (Figure \ref{fig:architecture}). Our architecture uses a ResNeXt-50 encoder \cite{resnext} to produce a feature vector $z$, which serves as the input to the pressure estimator, weak label estimator, and domain discriminator.

\subsubsection{Pressure Estimation}
Pressure is estimated using an FPN image-to-image network architecture \cite{fpn} which outputs a pressure image with the same dimensions as the input image, such that each pixel in the output corresponds to one pixel in the input. \network{} frames pressure estimation as a classification task where the continuous pressure space is divided into $b = 9$ discrete pressure bins, with the ground truth pressure bin $b_{gt}$. The network estimates the probability of each pressure bin, $\rho_{x,y}(b)$, for each pixel in the image. The pressure estimation task is supervised with a \textit{structure-aware cross-entropy} loss $L_p$ \cite{massa2016crafting, grady2023visual}. Unlike the standard cross-entropy loss, this loss penalizes small errors less than large errors.

\begin{align}
    L_p = -\sum_{x,y}\sum_{b}e^{-|b-b_{gt}|}log(\rho_{x,y}(b))
\end{align}

\begin{figure}[t]
  \centering
  \vspace{2mm}
  \includegraphics[width=.4\linewidth]{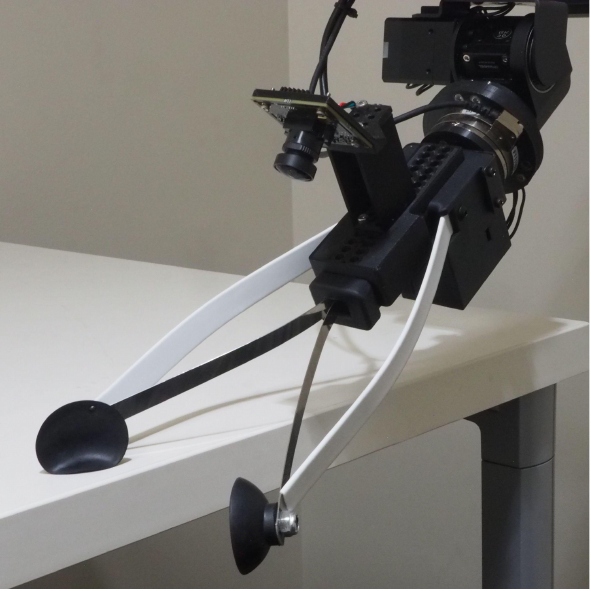}
  \vspace{-1mm}
  \caption{The gripper included on the Stretch RE1 robot features soft rubber fingertips supported by spring steel flexures. Both the fingertip and flexures deform under the application of force, resulting in visual cues that can be used to estimate contact pressure.}
  \label{fig:gripper_deflection}
\end{figure}

\subsubsection{Estimation of Forces and Torques}
\network{} performs the auxiliary task of estimating the forces and torques exerted by the gripper as measured by a wrist-mounted force/torque sensor. Unlike pressure data, which requires applying sensors to the objects being manipulated, force/torque measurements can be collected without altering the environment. We demonstrate that the inference of forces and torques informs the inference of pressures, and vice versa.



The feature vector $z$ from the image encoder is fed into a single convolutional layer followed by an MLP to estimate force/torque values. This is supervised by $L_{ft}$, a weighted sum of the L2 distances between the actual and estimated forces and torques.

\begin{align}
    L_{ft} = ||F - \hat{F}||^2 + c||T - \hat{T}||^2
\end{align}

Where $F, T$ are ground truth forces and torques, and $\hat{F}, \hat{T}$ are estimated forces and torques. Following \cite{vfts}, we choose $c$ to be the ratio of the force and torque standard deviations in the training dataset. This is done to account for the difference in scale between forces and torques.


\subsubsection{Adversarial Domain Adaptation}
We employ an unsupervised domain adaptation loss to further leverage data without pressure labels. Unlike natural environments, a pressure sensing array and fiducial markers appear in fully labeled images. We use a domain discriminator to align the distributions of features produced by fully labeled and weakly labeled images.

Our domain discriminator uses a classification head consisting of a convolutional layer followed by an MLP to classify whether an image came from the fully labeled or weakly labeled domain. Inspired by DANN \cite{ganin2016dann}, during training, the gradients from the discriminator loss are flipped before backpropagating through the image encoder. This trains the image encoder to produce features that maximally decrease the performance of the domain discriminator, thus making the image features invariant to the domain of the input. The discriminator is trained with loss $L_d$ to identify features from the fully labeled domain $z_f$ and weakly labeled domain $z_w$.


\begin{align}
    L_d = -log(D(z_f)) -log(1-D(z_w))
\end{align}

\subsubsection{Training Details}
We trained \network{} for a total of 200k iterations using the Adam \cite{adam} optimizer. The learning rate was set to 1e-3 for the first 150k iterations, and 1e-4 thereafter. During training, each batch included 14 fully labeled and 14 weakly labeled images. 

\network{} is trained with the weighted sum of the pressure loss, the force/torque loss, and the domain adaptation loss:

\begin{align}
    L = L_p + \lambda_1 L_{ft} + \lambda_2 L_d
\end{align}

We choose $\lambda_1$ to be 5e-3 and $\lambda_2$ to be 1e-3.

\begin{figure}[t]
  \centering
  \vspace{2mm}
  \includegraphics[width=1.0\linewidth]{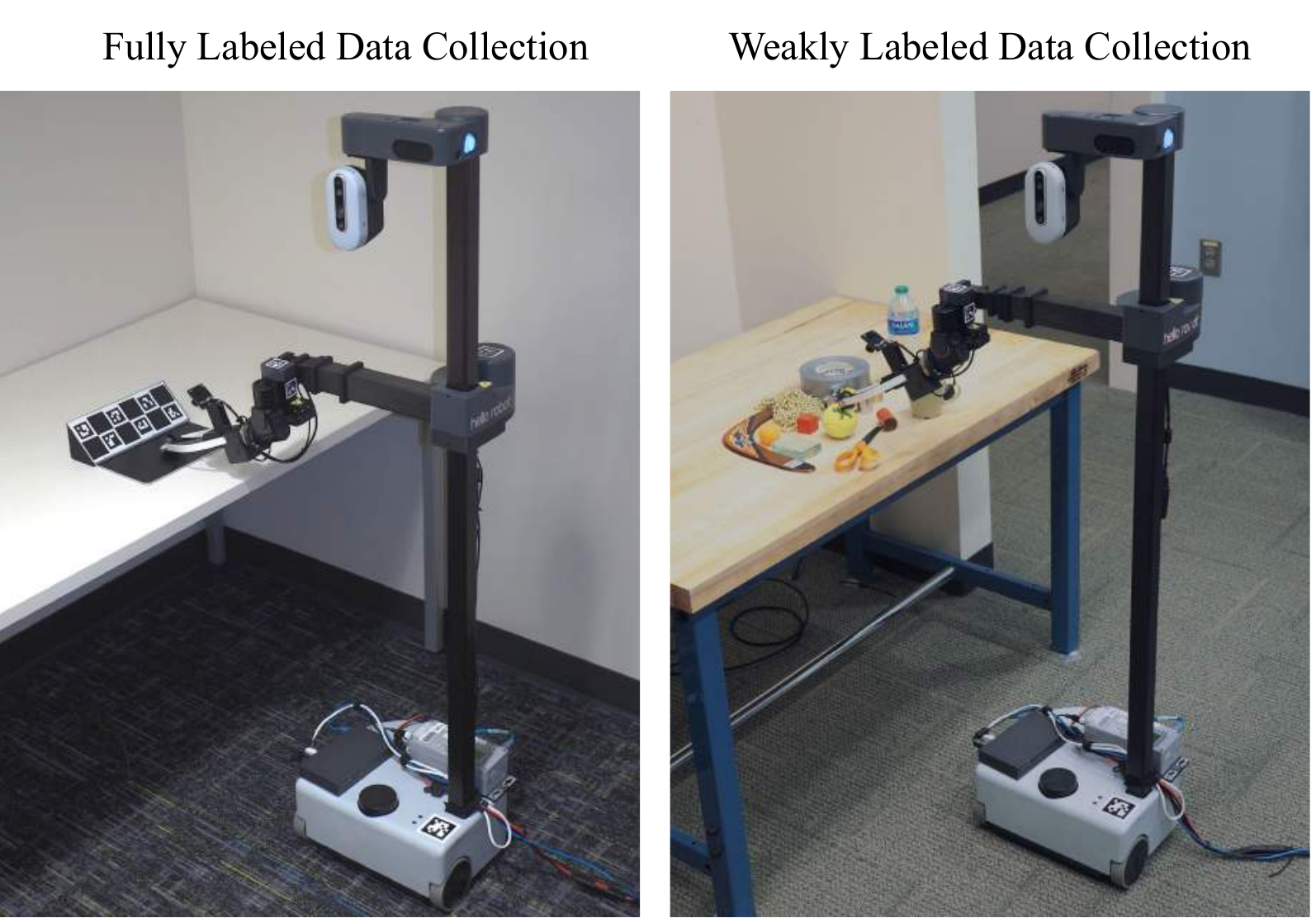}
  \vspace{-5mm}
  \caption{\textbf{Left:} Fully labeled data collection uses a pressure sensing array and a wrist-mounted force/torque sensor. \textbf{Right:} Weakly labeled data collection only uses a wrist-mounted force/torque sensor.}
  \label{fig:Weak_vs_Full}
\end{figure}

\subsection{Data Collection}

\subsubsection{Robotic Gripper}


For this paper, we use the Stretch RE1 mobile manipulator from Hello Robot. The robot includes a soft gripper with flexible tendon-actuated fingers and rubber fingertips \cite{kemp2021design}. Applying external forces to the gripper causes visible deformation of the fingertips and fingers (Figure \ref{fig:gripper_deflection}). Although our experiments focus solely on this gripper, we expect our methods could work with other soft grippers that visibly deform. For example, earlier methods have been successful with the Stretch RE1 gripper and a gripper with soft pneumatic fingers \cite{vpec, vfts}.


\subsubsection{Pressure Sensor}
To obtain ground truth pressure, we use a Sensel Morph, which is a flat, 23x13 cm pressure-sensing array consisting of a grid of force-sensitive resistor (FSR) sensors. 

\subsubsection{Force/Torque Sensor}
During data collection, we mount a 6-DoF ATI Mini45 force/torque sensor between the gripper and the wrist of the robot. Before each recording, we set the gripper to a level pose and zero the force/torque sensor readings, following \cite{vfts}. As a consequence, \network{} estimates the forces and torques applied to the wrist of the gripper, minus the gravitational load when the gripper is level.

We emphasize that the force/torque sensor is only used to collect training data. A trained network only uses an image as input.

\subsubsection{Camera}
We use a Sony IMX291 USB camera to obtain images of the gripper. The camera is mounted in an ``eye-in-hand" configuration (Figure \ref{fig:gripper_deflection}) such that it observes the gripper's flexures and fingertips, as well as the object being manipulated. By mounting the camera to the robot rather than the environment, we are able to capture images during manipulation in diverse environments.

Images are captured at 25 Hz and synchronized with force/torque readings. The camera's intrinsic parameters are used to rectify the image. For each frame containing a pressure sensor, the transform from the pressure sensor to the camera is obtained with a ChArUco fiducial board \cite{aruco, opencv_library} rigidly mounted to the sensor. Using this transform, ground truth pressure readings can be projected into the image (Figure \ref{fig:collage}).


\subsection{Data Capture Setup} \label{sec:data_capture}
When collecting the dataset, we record two types of data: fully labeled data and weakly labeled data (Figure \ref{fig:Weak_vs_Full}). Fully labeled data consists of RGB video frames paired with pressure and force/torque measurements. Weakly labeled data consists of RGB video frames paired with \textit{only} force/torque measurements.



\subsection{Data Capture Protocol}
We capture a training dataset with 3 hours of data in multiple lab and office environments. The robot is teleoperated during data collection to achieve contact with a variety of surfaces. We collect a total of 47k fully labeled images and 275k weakly labeled images. 

For fully labeled data, we adhere contact paper to the surface of the pressure sensor to emulate varied backgrounds. In the fully labeled training set, only solid color backgrounds are used in order to demonstrate generalization to the weakly labeled test set, which contains more challenging backgrounds. We also place distractor objects around the pressure sensor, but do not apply forces to the gripper or sensor with the distractor objects to avoid false measurements. 


When collecting weakly labeled data, the robot interacts with a wide variety of flat, curved, and compliant objects and surfaces. Many of the contact surfaces would be highly challenging to instrument with pressure sensors. 

\begin{figure}[t]
  \centering
  \vspace{2mm}
  \includegraphics[width=1\linewidth]{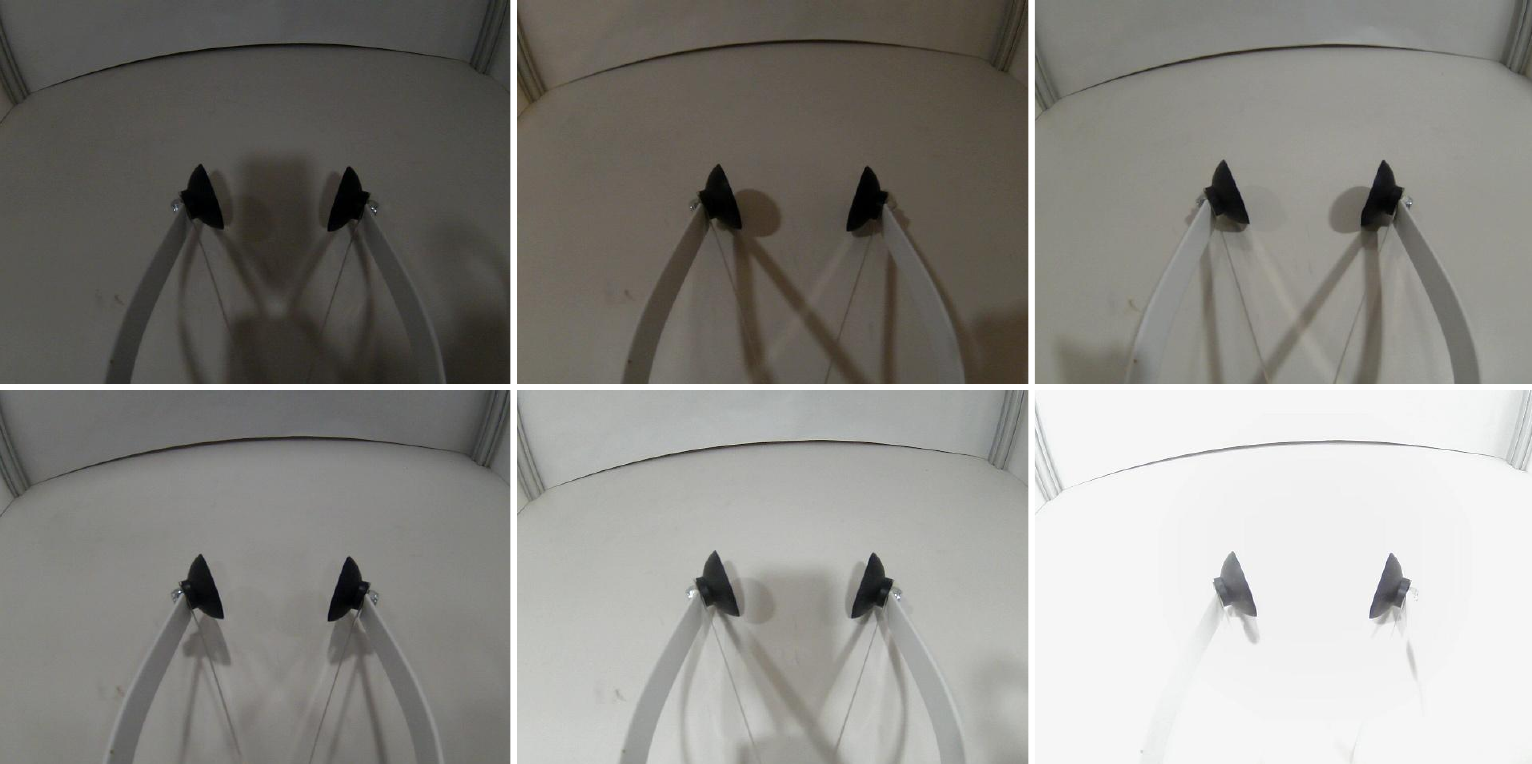}
  \vspace{-5mm}
  \caption{In pilot experiments, we found that our method is sensitive to changes in lighting. During collection of the training set, we use programmable smart lights to rapidly collect data for diverse lighting conditions. We additionally apply artificial brightness, contrast, saturation, and hue augmentation during training.}
  \label{fig:textures}
\end{figure}

During pilot experiments, we found that our networks are sensitive to changes in lighting. To increase the diversity of lighting, we developed a lighting randomization setup featuring four smart lightbulbs (Figure \ref{fig:textures}). These lightbulbs are mounted in the scene and are programmed to randomly change their brightness and color temperature to quickly capture many combinations of lighting variation. As our method only relies on individual RGB frames rather than videos, the lighting configuration can be changed rapidly while the robot is making contact to collect a wide variety of lighting scenarios.



We obtain 1 hour of testing data in a held-out household environment, including a bedroom, living room, and kitchen. We did not modify the lighting or objects present. Approximately half of the testing data was collected with a pressure sensor, on which held-out textures were adhered, including a marble, wood, and newspaper pattern. The remaining half of the testing data was collected while interacting with objects in their natural context within the environment.


\begin{figure*}[t]
  \centering
  \vspace{2mm}
  \includegraphics[width=.95\linewidth]{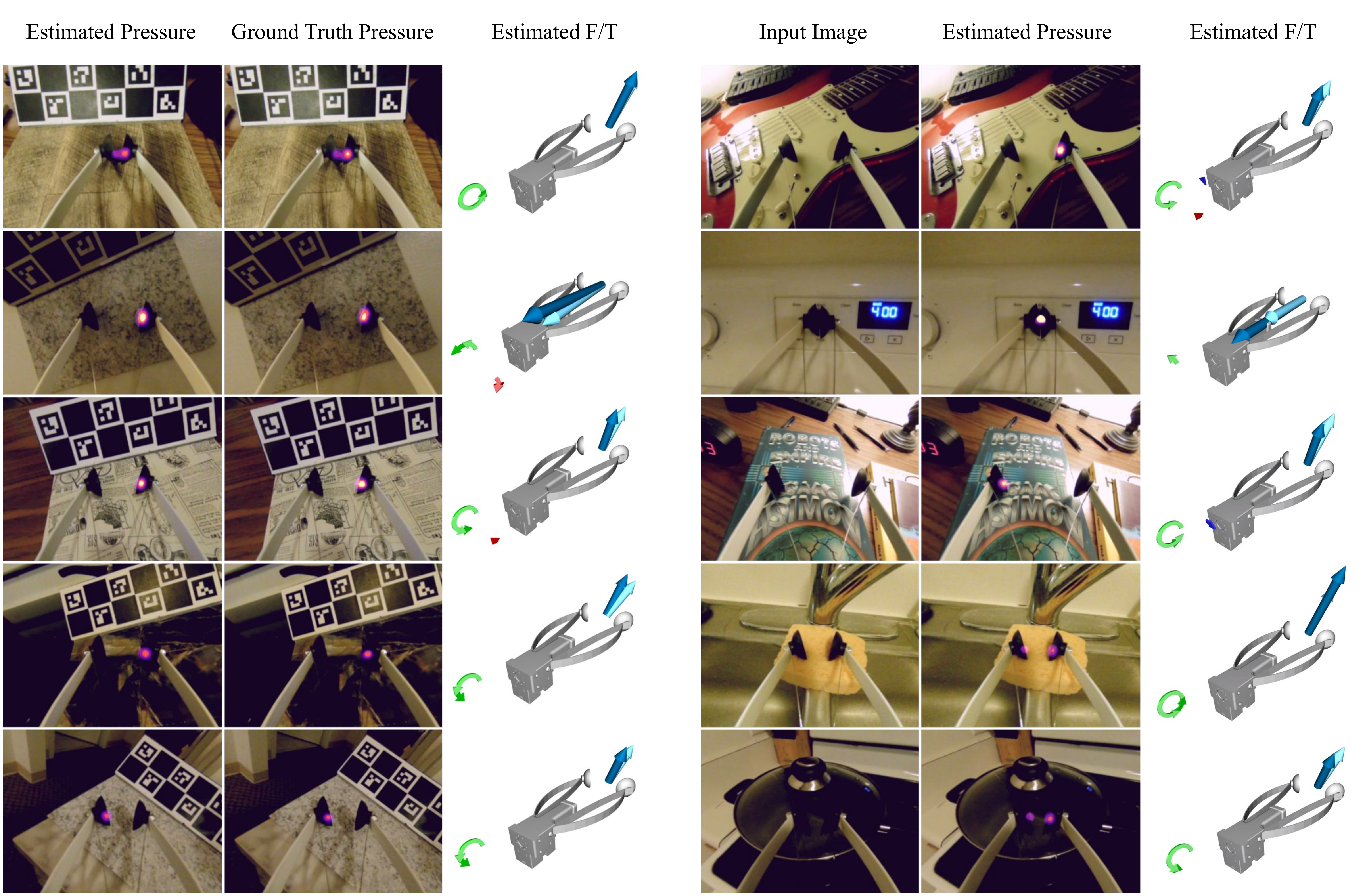}
  \vspace{-2mm}
  \caption{Results on the fully labeled (left) and weakly labeled (right) test sets. All objects and surfaces present in the test set are held out from both partitions of the training set. The force/torque visualization shows ground truth forces and torques as dark arrows, and estimated forces and torques as light arrows.} 
  
  \label{fig:collage}
\end{figure*}

\begin{table*}
\centering
\begin{tabular}{c|c|c||c|c|c}
      & \multicolumn{2}{c||}{\textbf{Fully Labeled Test Set}} & \multicolumn{3}{c}{\textbf{Weakly Labeled Test Set}} \\\hline
    \textbf{Method} & \textbf{Contact Acc.} & \textbf{Volumetric IoU} & \textbf{Contact Acc.} & \textbf{RMSE\boldmath$_F$} (N) & \textbf{RMSE\boldmath$_T$} (Nm) \\\hline
    VPEC \cite{vpec} & 83.9\% & 28.9\% & 84.2\% & - & - \\\hline
    VFTS \cite{vfts} & - & - & - & 2.07 & 0.28\ \\\hline
    \network{} (ours) & \textbf{92.1\%} & \textbf{40.3\%} & \textbf{92.9\%} & \textbf{1.57} & \textbf{0.19} \\\hline
\end{tabular}
\caption{Performance of \method{} as compared to VPEC and VFTS baselines.}
\label{tab:main_results}
\vspace{-1mm}
\end{table*}


\section{Evaluations}

We evaluate \network{} on a test set captured in a held-out home environment. \network{} is shown to outperform baselines from prior work. We also demonstrate the value of our contributions using ablations.

\subsection{Baselines}

We compare our approach to two prior works:

\subsubsection{VPEC \cite{vpec}} VPEC uses a camera mounted to the environment to capture images of a gripper interacting with a flat surface. It infers image-space pressure in a similar way to \method{}, however VPEC only uses fully labeled data. Due to the large difference in camera pose, we retrain VPEC on our dataset for comparison.

\subsubsection{VFTS \cite{vfts}}  While assessing the accuracy of force/torque estimates is not the focus of our paper, we still compare to VFTS, which estimates forces and torques applied to a gripper from images. VFTS trains a convolutional neural network to operate on images from an eye-in-hand camera. We also retrain this method on our dataset.

\subsection{Metrics}


We seek to evaluate \method{}'s pressure estimation performance on interactions with the natural world, but face the same issue that ground truth pressure is not available in these scenarios. As such, we evaluate on a fully labeled test set and a weakly labeled test set. More rigorous evaluations can be calculated on fully labeled data where ground truth pressure is available; however, evaluations on diverse, natural environments must be calculated on weakly labeled data.

We evaluate \network{}'s ability to estimate pressure, forces, and torques using several quantitative metrics following prior work \cite{vfts, grady2022pressurevision}.





\paragraph{Volumetric IoU}
To evaluate the accuracy of pressure estimates, we calculate the intersection over union between the estimated and ground truth pressure maps, $\hat{P}$ and $P$. The volumetric IoU views a 2D pressure image as a 3D pressure volume, and the height of the volume at that pixel $i, j$ is equal to the amount of pressure. The IoU is computed between the estimated and ground truth pressure volumes.

\begin{equation}
    IoU_{vol}=\frac{\sum^{i,j}min(P_{i,j}, \hat{P}_{i,j})}{\sum^{i,j}max(P_{i,j}, \hat{P}_{i,j})}
\end{equation}



\paragraph{Contact Accuracy}
We evaluate \network{} on each example in the fully labeled and weakly labeled datasets to determine its ability to infer the presence or absence of contact. We define a fully labeled example as being in contact if any pixel in $P$ exceeds a threshold of 1 kPa. For weakly labeled examples, we apply a 3N threshold to the force measurements from the force/torque sensor to determine if the gripper is in contact. For all examples, we say that contact is predicted if any pixel in $\hat{P}$ exceeds a threshold of 1 kPa. The contact accuracy is then defined as the percentage of examples for which contact is correctly predicted.

\newcommand{\y}{\textbf{\checkmark}}
\newcommand{\n}{$\mathbf{\times}$}
\begin{table*}
\centering
\begin{tabular}{c|c|c|c||c|c}
    \multicolumn{2}{c|}{\textbf{}} & \multicolumn{2}{c||}{\textbf{Fully-Labeled Test Set}} & \multicolumn{2}{c}{\textbf{Weakly-Labeled Test Set}} \\\hline
    \textbf{Domain Loss} & \textbf{Force/Torque Loss} & \textbf{Contact Acc.} & \textbf{Volumetric IoU} & \textbf{RMSE\boldmath$_F$} (N) & \textbf{RMSE\boldmath$_T$} (Nm) \\\hline
    & & 87.2\% & 21.1\% & - & - \\\hline
    \y & & 86.4\% & 27.2\% & - & - \\\hline
    & \y & 89.9\% & 37.4\% & 1.77 & 0.38 \\\hline
    \y & \y & \textbf{92.1\%} & \textbf{40.3\%} & \textbf{1.57} & \textbf{0.19} \\\hline
\end{tabular}
\caption{Ablation of Domain and Force/Torque losses of \network{}.}
\label{tab:data_ablations}
\vspace{-3mm}
\end{table*}

\begin{figure*}[t]
  \centering
  \vspace{4mm}
  \includegraphics[width=1\linewidth]{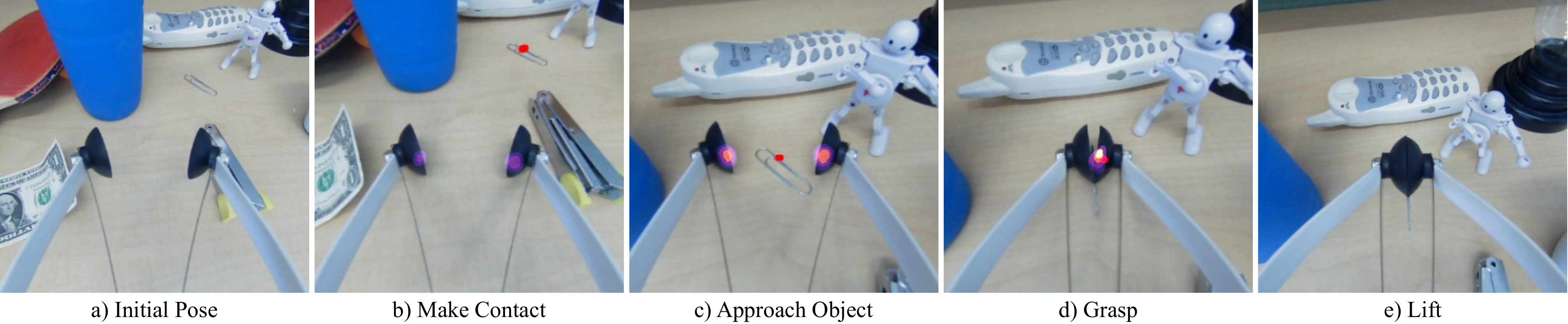}
  \caption{Example of a grasping sequence to pick up a paperclip in clutter. a) The robot begins above the surface. b) A user selects a point to grasp in the image, and the robot moves to make contact with the surface. c) The average of the estimated fingertip positions is used to visually servo the fingertips and encircle the object. d-e) The gripper closes around the object and lifts.}
  \label{fig:grasping_task}
  \vspace{-4mm}
\end{figure*}

\begin{figure}[t]
  \centering
  \vspace{2mm}
  \includegraphics[width=.9\linewidth]{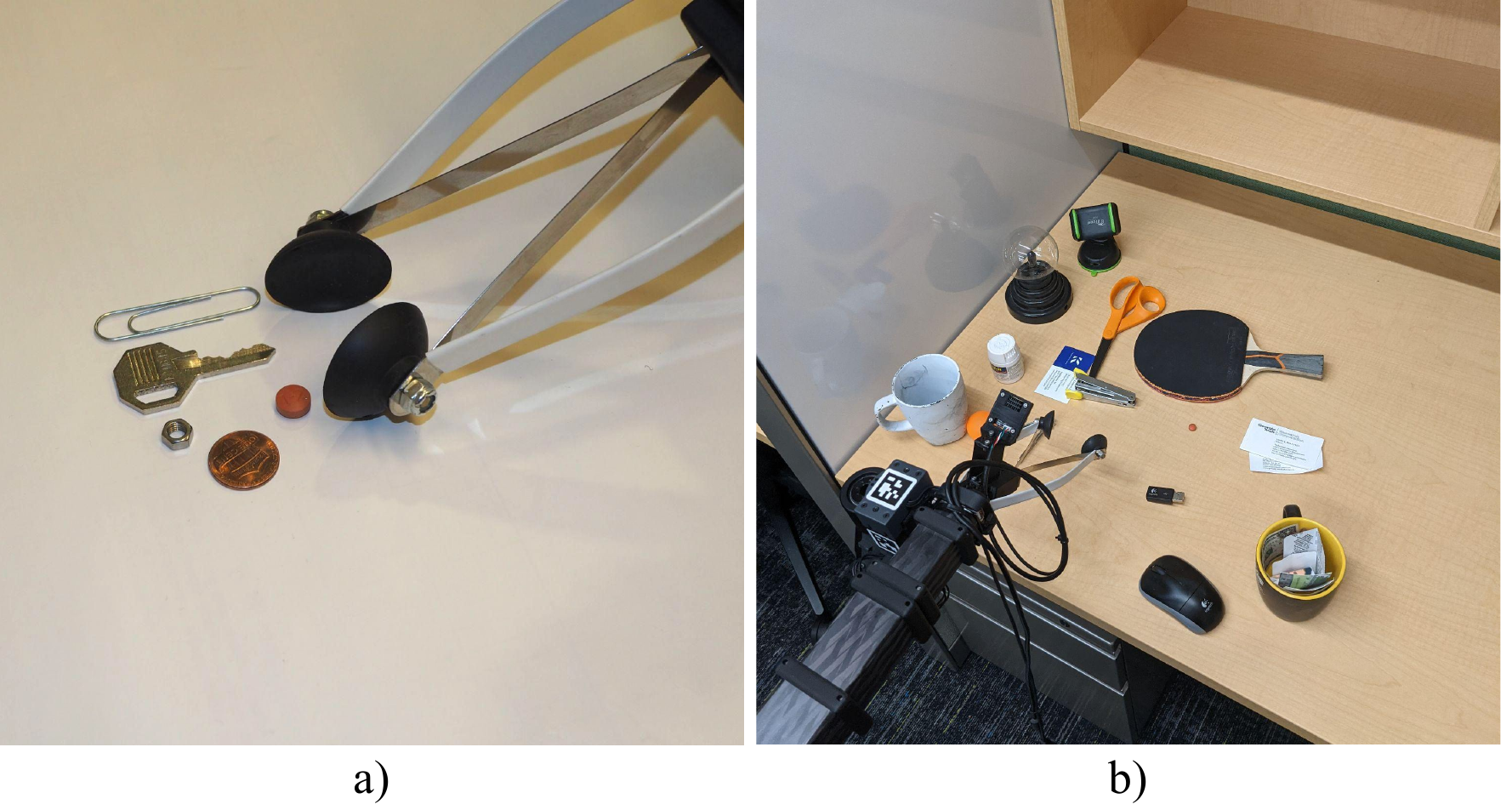}
  \vspace{-2mm}
  \caption{a) The objects used in the grasping from clutter task included a paperclip, a key, a nut, a penny, and a pill. b) Setup for the grasping from clutter task in one of the 5 unseen environments.}
  \label{fig:tasks_deflection}
  \vspace{-5mm}
\end{figure}

\paragraph{Force and Torque RMSE}
We calculate the root-mean-square error between the estimated and measured forces and torques.


\subsection{Results}

We evaluate \network{} on the fully labeled and weakly labeled test set, shown in Table \ref{tab:main_results}.  We find that our approach can estimate pressure in diverse environments, and show examples in Figure \ref{fig:collage}.

\method{} outperforms VPEC on all metrics. As VPEC cannot utilize diverse weakly labeled data, it performs worse on a test set collected in a real home.

We also compare our approach to VFTS, a method designed specifically to estimate forces and torques from images. Our method outputs both force/torque and pressure, and achieves lower error than VFTS.

We expect that forces and torques inform contact pressure prediction, and vice versa, due to their physical relationship, which is grounded in mechanics. For example, if we disregard shear forces, the resultant force applied by the gripper to the force/torque sensor should be equal to the integral of the pressure distribution applied to the gripper. It then follows that low contact pressure should result in low-magnitude forces, and high contact pressure should result in high-magnitude forces.

\subsection{Ablations}

Table \ref{tab:data_ablations}
shows results when ablating two components of the network, the domain discriminator and the force/torque estimator. When neither of these components is used, \network{} is trained on only the fully labeled data to classify the pressure for each pixel in the input image, and does not use the weakly labeled data in training. The domain loss leverages the additional data in an unsupervised fashion, and improves volumetric IoU by 29\%. The force/torque loss provides a stronger benefit, improving volumetric IoU by 77\%. The two losses combined improve volumetric IoU by 91\%.

\section{Pressure-Based Visuomotor Control} \label{sec:robotic_system}

\subsection{Grasping from Clutter in the Wild} \label{sec:grasp}
Robotic manipulation in unstructured environments remains a challenging problem. One prevalent manipulation objective is \textit{grasping from clutter}, in which a robot is tasked with picking up a desired object from an arrangement of distractor objects and other obstacles. This objective is relevant to many real-world applications, including assistive household tasks. 


To demonstrate the effectiveness of \method{}, we design a grasping algorithm to autonomously grasp small objects, inspired by \cite{vpec}. To reliably grasp small objects such as a paperclip, our grasping strategy slides the fingertips across a table's surface to encircle the object, and then the gripper closes. However, when sliding, the gripper deflects by a large amount (Figure \ref{fig:gripper_deflection}), such that it is difficult to control the precise fingertip position open-loop. Additionally, the force with which the gripper presses against the surface must be regulated carefully so that it slides across the surface while still making sufficient contact to grasp thin objects. 


We design a closed-loop controller which uses \network{}'s pressure estimates to achieve this strategy. The location of fingertip pressure is moved relative to a target, and the magnitude of the pressure is regulated for sliding and grasping (Figure \ref{fig:grasping_task}). 



During each trial, the gripper begins above an uncluttered portion of a flat surface approximately 50 cm away from the target object. A human operator then clicks on the desired object in an image captured by the robot's camera via a user interface. Next, the robot autonomously lowers towards the surface until \network{} estimates pressure above a specified threshold. This pressure $\hat{P}$ is used to localize the fingertips in the image. An error vector $\epsilon_i$ is then computed by taking the difference between the center of pressure and the object location, which is tracked in the image with a filter \cite{lukezic2017discriminative}. This error is then projected from image space to the robot's task space to produce a task space error vector $\epsilon_t$. At each time step, the robot is commanded to move in the direction $\epsilon_t$ toward the target location. A simple controller simultaneously maintains the desired sliding pressure using the magnitude of $\hat{P}$. Once the fingertip pressure location is aligned with the object, the gripper is closed.

The robot grasps a total of 5 unseen, low-profile objects, each for 10 trials, in 5 novel, cluttered environments. These environments contained surfaces not present in any of the fully or weakly labeled training, with the exception of a glass surface which was present in the weakly labeled training set, but was placed above an unseen surface. Each object is randomly positioned and oriented between trials, and the distractor objects are rearranged. Since the tracker was not the focus of our research, we excluded the two trials that failed due to tracking errors.

\subsubsection{Results}
 Our results show that our pressure-based visuomotor controller can effectively grasp objects in unstructured, cluttered environments, demonstrating the potential of our method for real-world applications. Given only images from a gripper-mounted camera, our system was able to grasp all 5 test objects from 5 cluttered, previously unseen environments, producing an overall success rate of 90\%. While widely applicable, \method{} has limitations. For example, it is not robust to occlusion of the fingertips, is less reliable when operating on backgrounds matching the color of the fingertips, and fails when subjected to loads not represented in the training data, such as grip force and forces applied at locations other than the fingertips.

\begin{table}
\centering
\begin{tabular}{c|c|c}
    \textbf{Object} & \textbf{Dims. L$\times$W$\times$H} & \textbf{Successes/Trials}\\\hline
    Paperclip & 50$\times$10$\times$1 mm  & 9/10  \\\hline
    Penny & 19$\times$19$\times$1 mm  & 8/10  \\\hline
    Nut & 8$\times$8$\times$4 mm  & 10/10 \\\hline
    Pill & 10$\times$10$\times$5 mm  & 10/10  \\\hline
    Key & 45$\times$22$\times$2 mm  & 8/10 \\\hline
\end{tabular}
\caption{Grasping Results}
\label{tab:grasping_results}
\vspace{-10mm}
\end{table}

\section{Conclusion}

We introduced \method{}, a method to visually estimate the contact pressure applied by a soft robotic gripper to the environment. Obtaining ground truth pressure measurements for diverse environments is challenging. In contrast, ground truth force/torque measurements are relatively easy to obtain. By using force/torque measurements as weak labels, \method{} outperforms prior methods, enables precision manipulation in cluttered settings, and provides accurate estimates for unseen conditions relevant to in-home use.




\bibliographystyle{IEEEtran}
\bibliography{cited}

\begin{thebibliography}{10}
\providecommand{\url}[1]{#1}
\csname url@rmstyle\endcsname
\providecommand{\newblock}{\relax}
\providecommand{\bibinfo}[2]{#2}
\providecommand\BIBentrySTDinterwordspacing{\spaceskip=0pt\relax}
\providecommand\BIBentryALTinterwordstretchfactor{4}
\providecommand\BIBentryALTinterwordspacing{\spaceskip=\fontdimen2\font plus
\BIBentryALTinterwordstretchfactor\fontdimen3\font minus
  \fontdimen4\font\relax}
\providecommand\BIBforeignlanguage[2]{{%
\expandafter\ifx\csname l@#1\endcsname\relax
\typeout{** WARNING: IEEEtran.bst: No hyphenation pattern has been}%
\typeout{** loaded for the language `#1'. Using the pattern for}%
\typeout{** the default language instead.}%
\else
\language=\csname l@#1\endcsname
\fi
#2}}

\bibitem{vpec}
P.~Grady, J.~A. Collins, S.~Brahmbhatt, C.~D. Twigg, C.~Tang, J.~Hays, and
  C.~C. Kemp, ``Visual pressure estimation and control for soft robotic
  grippers,'' \emph{2022 IEEE/RSJ International Conference on Intelligent
  Robots and Systems (IROS)}, 2022.

\bibitem{nazari2021image}
A.~A. Nazari, F.~Janabi-Sharifi, and K.~Zareinia, ``Image-based force
  estimation in medical applications: A review,'' \emph{IEEE Sensors Journal},
  vol.~21, no.~7, pp. 8805--8830, 2021.

\bibitem{kennedy2005vision}
C.~W. Kennedy and J.~P. Desai, ``A vision-based approach for estimating contact
  forces: Applications to robot-assisted surgery,'' \emph{Applied Bionics and
  Biomechanics}, vol.~2, no.~1, pp. 53--60, 2005.

\bibitem{noohi2014using}
E.~Noohi, S.~Parastegari, and M.~{\v{Z}}efran, ``Using monocular images to
  estimate interaction forces during minimally invasive surgery,'' in
  \emph{2014 IEEE/RSJ International Conference on Intelligent Robots and
  Systems}.\hskip 1em plus 0.5em minus 0.4em\relax IEEE, 2014, pp. 4297--4302.

\bibitem{marban2019recurrent}
A.~Marban, V.~Srinivasan, W.~Samek, J.~Fern{\'a}ndez, and A.~Casals, ``A
  recurrent convolutional neural network approach for sensorless force
  estimation in robotic surgery,'' \emph{Biomedical Signal Processing and
  Control}, vol.~50, pp. 134--150, 2019.

\bibitem{chua2021toward}
Z.~Chua, A.~M. Jarc, and A.~M. Okamura, ``Toward force estimation in
  robot-assisted surgery using deep learning with vision and robot state,'' in
  \emph{2021 IEEE International Conference on Robotics and Automation
  (ICRA)}.\hskip 1em plus 0.5em minus 0.4em\relax IEEE, 2021, pp.
  12\,335--12\,341.

\bibitem{kim2019efficient}
D.~Kim, H.~Cho, H.~Shin, S.-C. Lim, and W.~Hwang, ``An efficient
  three-dimensional convolutional neural network for inferring physical
  interaction force from video,'' \emph{Sensors}, vol.~19, no.~16, p. 3579,
  2019.

\bibitem{erickson_visual_haptic_reasoning_2022}
\BIBentryALTinterwordspacing
Y.~Wang, D.~Held, and Z.~Erickson, ``Visual haptic reasoning: Estimating
  contact forces by observing deformable object interactions,'' \emph{{IEEE}
  Robotics Autom. Lett.}, vol.~7, no.~4, pp. 11\,426--11\,433, 2022. [Online].
  Available: \url{https://doi.org/10.1109/LRA.2022.3199684}
\BIBentrySTDinterwordspacing

\bibitem{kuppuswamy2020soft}
N.~Kuppuswamy, A.~Alspach, A.~Uttamchandani, S.~Creasey, T.~Ikeda, and
  R.~Tedrake, ``Soft-bubble grippers for robust and perceptive manipulation,''
  in \emph{2020 IEEE/RSJ International Conference on Intelligent Robots and
  Systems (IROS)}.\hskip 1em plus 0.5em minus 0.4em\relax IEEE, 2020, pp.
  9917--9924.

\bibitem{ward2018tactip}
B.~Ward-Cherrier, N.~Pestell, L.~Cramphorn, B.~Winstone, M.~E. Giannaccini,
  J.~Rossiter, and N.~F. Lepora, ``The tactip family: Soft optical tactile
  sensors with 3d-printed biomimetic morphologies,'' \emph{Soft Robotics},
  vol.~5, no.~2, pp. 216--227, 2018.

\bibitem{yamaguchi2016combining}
A.~Yamaguchi and C.~G. Atkeson, ``Combining finger vision and optical tactile
  sensing: Reducing and handling errors while cutting vegetables,'' in
  \emph{2016 IEEE-RAS 16th International Conference on Humanoid Robots}.\hskip
  1em plus 0.5em minus 0.4em\relax IEEE, 2016, pp. 1045--1051.

\bibitem{yuan2017gelsight}
W.~Yuan, S.~Dong, and E.~H. Adelson, ``Gelsight: High-resolution robot tactile
  sensors for estimating geometry and force,'' \emph{Sensors}, vol.~17, no.~12,
  p. 2762, 2017.

\bibitem{lambeta2020digit}
M.~Lambeta, P.-W. Chou, S.~Tian, B.~Yang, B.~Maloon, V.~R. Most, D.~Stroud,
  R.~Santos, A.~Byagowi, G.~Kammerer, \emph{et~al.}, ``Digit: A novel design
  for a low-cost compact high-resolution tactile sensor with application to
  in-hand manipulation,'' \emph{IEEE Robotics and Automation Letters}, vol.~5,
  no.~3, pp. 3838--3845, 2020.

\bibitem{vfts}
J.~A. Collins, P.~Grady, and C.~C. Kemp, ``Force/torque sensing for soft
  grippers using an external camera,'' \emph{IEEE International Conference on
  Robotics and Automation (ICRA)}, 2023.

\bibitem{grady2023visual}
\BIBentryALTinterwordspacing
P.~Grady, J.~A. Collins, C.~Tang, C.~D. Twigg, J.~Hays, and C.~C. Kemp,
  ``Visual estimation of fingertip pressure on diverse surfaces using easily
  captured data,'' 2023. [Online]. Available:
  \url{https://arxiv.org/abs/2301.02310}
\BIBentrySTDinterwordspacing

\bibitem{long2015dan}
M.~Long, Y.~Cao, J.~Wang, and M.~Jordan, ``Learning transferable features with
  deep adaptation networks,'' in \emph{International conference on machine
  learning}.\hskip 1em plus 0.5em minus 0.4em\relax PMLR, 2015, pp. 97--105.

\bibitem{tzeng2014deepdomainconfusion}
E.~Tzeng, J.~Hoffman, N.~Zhang, K.~Saenko, and T.~Darrell, ``Deep domain
  confusion: Maximizing for domain invariance,'' \emph{arXiv preprint
  arXiv:1412.3474}, 2014.

\bibitem{long2016unsupervised}
M.~Long, H.~Zhu, J.~Wang, and M.~I. Jordan, ``Unsupervised domain adaptation
  with residual transfer networks,'' \emph{Advances in neural information
  processing systems}, vol.~29, 2016.

\bibitem{ganin2016dann}
Y.~Ganin, E.~Ustinova, H.~Ajakan, P.~Germain, H.~Larochelle, F.~Laviolette,
  M.~Marchand, and V.~Lempitsky, ``Domain-adversarial training of neural
  networks,'' \emph{The journal of machine learning research}, vol.~17, no.~1,
  pp. 2096--2030, 2016.

\bibitem{ahn2018learning}
J.~Ahn and S.~Kwak, ``Learning pixel-level semantic affinity with image-level
  supervision for weakly supervised semantic segmentation,'' in
  \emph{Proceedings of the IEEE conference on computer vision and pattern
  recognition}, 2018, pp. 4981--4990.

\bibitem{chang2020weakly}
Y.-T. Chang, Q.~Wang, W.-C. Hung, R.~Piramuthu, Y.-H. Tsai, and M.-H. Yang,
  ``Weakly-supervised semantic segmentation via sub-category exploration,'' in
  \emph{Proceedings of the IEEE/CVF Conference on Computer Vision and Pattern
  Recognition}, 2020, pp. 8991--9000.

\bibitem{paul2020domain}
S.~Paul, Y.-H. Tsai, S.~Schulter, A.~K. Roy-Chowdhury, and M.~Chandraker,
  ``Domain adaptive semantic segmentation using weak labels,'' in
  \emph{European conference on computer vision}.\hskip 1em plus 0.5em minus
  0.4em\relax Springer, 2020, pp. 571--587.

\bibitem{resnext}
S.~Xie, R.~B. Girshick, P.~Doll{\'{a}}r, Z.~Tu, and K.~He, ``Aggregated
  residual transformations for deep neural networks,'' in \emph{2017 {IEEE}
  Conference on Computer Vision and Pattern Recognition, (CVPR)}, 2017.

\bibitem{fpn}
T.-Y. Lin, P.~Doll{\'a}r, R.~Girshick, K.~He, B.~Hariharan, and S.~Belongie,
  ``Feature pyramid networks for object detection,'' in \emph{IEEE Conference
  on Computer Vision and Pattern Recognition, (CVPR)}, 2017, pp. 2117--2125.

\bibitem{massa2016crafting}
F.~Massa, R.~Marlet, and M.~Aubry, ``Crafting a multi-task {CNN} for viewpoint
  estimation,'' \emph{BVMC}, 2016.

\bibitem{adam}
D.~P. Kingma and J.~Ba, ``Adam: {A} method for stochastic optimization,'' in
  \emph{3rd International Conference on Learning Representations, (ICLR) 2015},
  Y.~Bengio and Y.~LeCun, Eds., 2015.

\bibitem{kemp2021design}
C.~C. Kemp, A.~Edsinger, H.~M. Clever, and B.~Matulevich, ``The design of
  {S}tretch: {A} compact, lightweight mobile manipulator for indoor human
  environments,'' in \emph{IEEE International Conference on Robotics and
  Automation (ICRA)}, 2022.

\bibitem{aruco}
S.~Garrido-Jurado, R.~Mu{\~n}oz-Salinas, F.~J. Madrid-Cuevas, and M.~J.
  Mar{\'\i}n-Jim{\'e}nez, ``Automatic generation and detection of highly
  reliable fiducial markers under occlusion,'' \emph{Pattern Recognition},
  vol.~47, no.~6, pp. 2280--2292, 2014.

\bibitem{opencv_library}
G.~Bradski, ``{The OpenCV Library},'' \emph{Dr. Dobb's Journal of Software
  Tools}, 2000.

\bibitem{grady2022pressurevision}
P.~Grady, C.~Tang, S.~Brahmbhatt, C.~D. Twigg, C.~Wan, J.~Hays, and C.~C. Kemp,
  ``{PressureVision:} estimating hand pressure from a single {RGB} image,''
  \emph{European Conference on Computer Vision (ECCV)}, 2022.

\bibitem{lukezic2017discriminative}
A.~Lukezic, T.~Vojir, L.~ˇCehovin~Zajc, J.~Matas, and M.~Kristan,
  ``Discriminative correlation filter with channel and spatial reliability,''
  in \emph{Proceedings of the IEEE conference on computer vision and pattern
  recognition}, 2017, pp. 6309--6318.

\end{thebibliography}

\end{document}